# DRAFT: MACHINE LEARNING-BASED MULTI-OBJECTIVE DESIGN EXPLORATION OF FLEXIBLE DISC ELEMENTS

**Gehendra Sharma[1,*], Sungkwang Mun[1], Nayeon Lee[1], Luke Peterson[1], Daniela Tellkamp[1], and Anand Balu Nellippallil[2]**

[1]Center for Advanced Vehicular Systems
Mississippi State University, Starkville, MS, USA 39759

[2]Department of Mechanical and Civil Engineering
Florida Institute of Technology, Melbourne, FL, USA 32901

## ABSTRACT

*Design exploration is an important step in the engineering design process. This involves the search for design/s that meet the specified design criteria and accomplishes the predefined objective/s. In recent years, machine learning-based approaches have been widely used in engineering design problems. This paper showcases Artificial Neural Network (ANN) architecture applied to an engineering design problem to explore and identify improved design solutions. The case problem of this study is the design of flexible disc elements used in disc couplings. We are required to improve the design of the disc elements by lowering the mass and stress without lowering the torque transmission and misalignment capability. To accomplish this objective, we employ ANN coupled with genetic algorithm in the design exploration step to identify designs that meet the specified criteria (torque and misalignment) while having minimum mass and stress. The results are comparable to the optimized results obtained from the traditional response surface method. This can have huge advantage when we are evaluating conceptual designs against multiple conflicting requirements.*

## NOMENCLATURE

| | |
|---|---|
| ANN | Artificial Neural Network |
| RSM | Response Surface Model |
| GA | Genetic Algorithm |
| FEA | Finite Element Analysis |
| ML | Machine Learning |

## 1. PROBLEM STATEMENT: DESIGN OF FLEXIBLE DISC ELEMENTS

Couplings are mechanical components that join two rotating parts to transmit mechanical power. While transmitting mechanical power, it is also required to offer torque resilience to resist torsional forces caused by two rotating parts [1]. The inability to handle misalignment between the rotating parts and system loads causes early torsional failures [2]. A disc coupling is one kind of coupling that uses flexible disc elements in its design. The disc couplings can transmit high torque, operate at high speed, and compensate for misalignment than other designs do by using flexible disc elements [3]. The disc elements as shown in Figure 1 are generally stacked together and connected between rotating parts. As the disc elements play a role to provide torque and take misalignment, the design of these flexible disc elements is critical in designing a disc coupling.

Despite being one of the important transmission components, the lack of published resources limit designers' ability to make contribution in the improvement of disc design. In this paper, we detail the challenges and considerations to be taken for designing disc elements. We utilize these knowledge in formulating a design problem aimed towards improving disc design. We showcase the design improved in using two approaches, that are, coupled ANN-GA and coupled RSM-GA.

---

[*] Corresponding author [gs1092@msstate.edu]

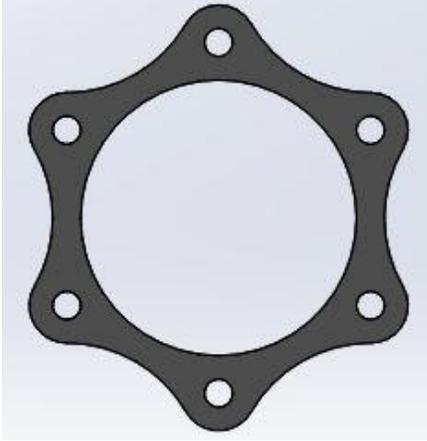

Figure 1: Flexible Disc Element

**2.1 Torque Transmission through Discs**

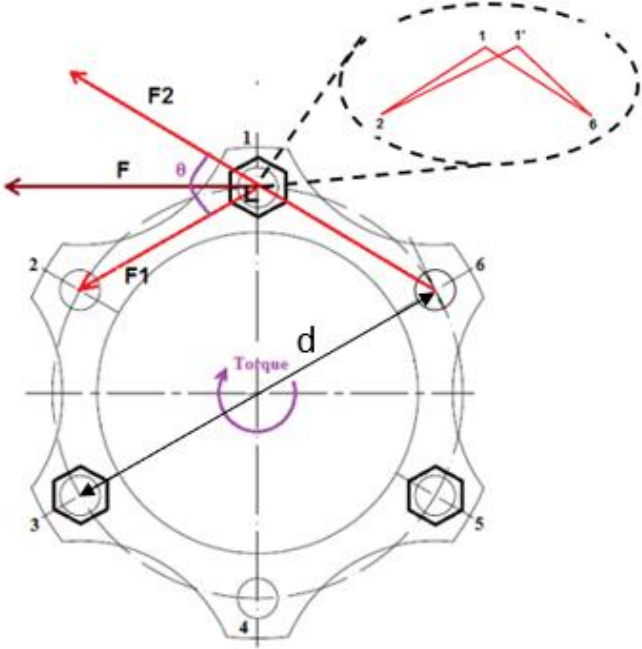

Figure 2: Disc Load During Torque Transmission

In Figure 2 is shown a disc design with 6 links. Each hole location labeled from 1 to 6 in the Figure 2 is referred to as joints. The segment between the consecutive joints is referred to as links. For example, the segment joining Joint 1 and Joint 6 is called Link 1_6 and so on. In detail, the discs are assembled in couplings in such a way that Joints 1,3 and, 5 are connected to one side of the rotating part while Joints 2, 4 and, 6 are connected to the other side (Figure 2). When the torque transmission takes place through these discs, torsional forces exerted to the discs cause deformation, as shown in Figure 2, resulting in the Joint 1 to shift to 1'. It should be noted that this is a very small shift. As a reaction to this shift, reaction forces are created in the Link 1_2 and Link 1_6. Link 1_6 is under compressive force while Link 1_2 is under tensile force. In Figure 2, F2 represents the reaction offered by Link 1_6 and F1 represents the reaction offered by Link 1_2. As these discs are thin structures, compressive force induces buckling in the disc links. Therefore, buckling load that the Link 1_6 can take limits the maximum reaction force F2 that the link can offer. It is to be noted that the discs are to be designed to have a better buckling resistance to improve the torque carrying capability as the buckling is one of the critical forces causing failure in the discs [4]. Another way to improve torque transmission capability is adding more discs. While adding more discs improves torque transmission capability, we restrict the scope of this work to improvement of the disc design.

As a result of the minute shift from of Joint 1 from 1 to 1', one link gets stretched and the other link gets compressed.

Elongation of link 1_2 = Length of 1'_2 – Length of 1_2

Compression of link 1_6 = Length of 1_6 – Length of 1'_6

As the shift is very small, these are nearly equal; hence, F1 = F2. As the disc has 6 links, $\Theta = 60^0$ and hence, we can establish that the resultant force (F) at joint 1 is

$$F = \sqrt{F1^2 + F2^2 + 2(F1)(F2)cos\Theta} \qquad \text{Equation 1}$$

Substituting $\Theta$ in Equation 1, we get,

$$F = \sqrt{3}\ F2 \qquad \text{Equation 2}$$

The disc has 3 links that undergo buckling and 3 links that undergo tensile stretch. Hence, the force resolution shown in Equation 2 can also be resolved at Joint 3 and Joint 5. As the disc has Pitch Circle Diameter (PCD) equal to d, the maximum torque (T) that the disc can transmit without buckling is given by:

$$T = (Number\ of\ buckling\ links) * F * \frac{d}{2} \qquad \text{Equation 3}$$

$$T = 3\sqrt{3}F2 * \frac{d}{2} \qquad \text{Equation 4}$$

As discussed previously, buckling will cause these thin structures to fail during torque transmission (assuming there are no misalignment). By improving the disc design to take more buckling load, we can enhance the torque transmission capability.

**2.2 Reaction Forces Due to Axial Misalignment in Discs**

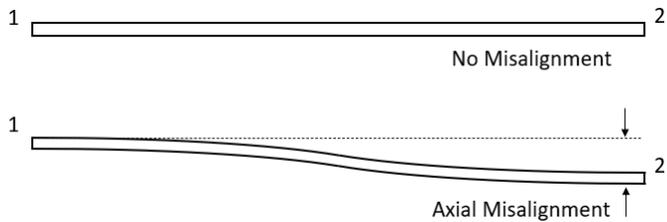

Figure 3: Deformation of the Link Due to Axial Misalignment in Discs

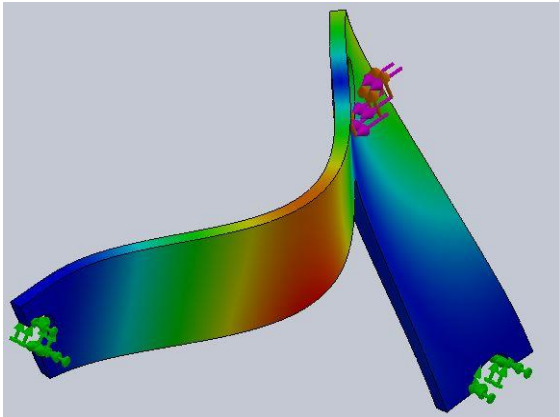

Figure 4: Buckling in Discs

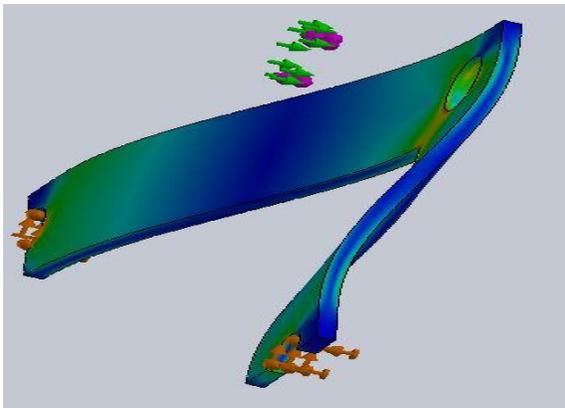

Figure 5: Bending in Discs

The axial misalignment causes bending in the discs as shown in Figure 3. This deformation causes high flexural stresses at the ends and are the reason that the discs fail. The stresses induced by torque transmission and axial misalignment are almost steady however tilting movement induces fluctuating stresses [4]. In this paper, we have considered torque and axial misalignment and hence focus on steady stresses. However, tilting in the discs can be added into the formulation by taking appropriate safety factors against fatigue failure.

There are two fundamental sources of loads/stresses in discs, i.e., torque and misalignment. Primarily, the torque induces buckling in the disc links (Figure 4), while the misalignment induces high bending stresses (Figure 5). While avoiding failure due to buckling and bending, we are interested in a disc design that has minimal mass and stress while being able to take specified axial misalignment and buckling load.

Due to the disc radial symmetry, Finite Element Analysis (FEA) is carried out in one-third disc segment to save computational time. Two different disc designs as shown in Figure 6 and Figure 7 are selected as base designs that need to be improved.

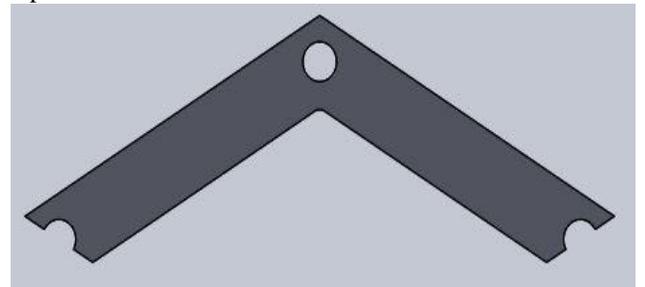

Figure 6: Design A

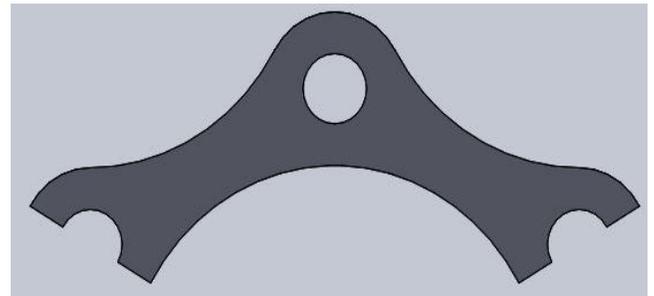

Figure 7: Design B

The design of flexible disc elements (Design A and Design B) that is to be used in disc couplings, is required. The disc has fixed number of links, that is, 6 (see Figure 1). The objective is to improve the quality of design disc while ensuring an ability to tolerate axial misalignment of 0.25 mm. The quality of design is measured in terms of design objectives. Specifically, we need a design that has a lower mass and stress without any sign of buckling in the discs due to torque load. Alloy steel with the following material properties (Table 1) is used in the design.

Table 1: Considered Material Properties of Alloy Steel for Disc Design A and B

| Material Properties | Values |
|---|---|
| Elastic Modulus | 210 GPa |
| Density | 7700 Kg/m3 |
| Yield Strength | 620 MPa |
| Poisson's Ratio | 0.28 |

## 2. INTRODUCTION

In this section, we present a review of machine learning-based methods applied to engineering design problems. Consequently, we present brief introduction to the deep learning techniques and approaches that give context to the works presented in this paper is provided. It includes a brief summary of the neural network architecture and learning strategies.

The availability of data and computational power and new methods have resulted in ML being applied to engineering design problems in several domains, such as materials design [5, 6], manufacturing design [7], topology optimization [8], etc. A novel deep learning-based method to carry out the optimal design without an iterative scheme is proposed by Yu and co-authors [9]. A framework to generate new structural and topology designs in an iterative fashion using generative adversarial networks is proposed by Oh and co-authors [10]. An ML-based method for real-time structural design optimization is proposed by Lei and co-authors [11]. Yang and co-authors used Generative adversarial networks to generate synthetic images of material microstructures [12]. Liu and Wang [13] proposed a multi-fidelity physics-constrained neural network (MF-PCNN) for material modeling. A parametric level set method for topology optimization based on a deep neural network is proposed by Deng and Albert [8]. All these applications demonstrate the success of ML tools and methods to address engineering problems across wide applications.

Deep learning is a powerful tool capable of processing a huge amount of unstructured information to generate insightful results. These impressive results come from the nonlinear processing of data in multiple layers [14]. Deep learning models have various architectures and can perform supervised, semi-supervised, and unsupervised learning [15]. The work presented in this paper uses Artificial Neural Networks (ANN) to carry out supervised learning on simulation data. Specifically, besides input and output layers, the networks include one or more intermediate layers between the input and output layers called hidden layers. A neuron or node in hidden layers can have one or more inputs of $x_i$, which come from the neurons in the input layer or a previous layer. These inputs are then multiplied with weights $w_i$ to be summed and shifted by a bias $b$, resulting in an intermediate single-valued result to pass through an activation function $f(\cdot)$ that smoothly maps the intermediate result in the desired range, e.g., 0 to 1 or -1 to 1. This sequence of calculations, multiplication-sum-shift-mapping, generates a resulting output $h$ and is performed as many as a predefined number of neurons in the next layer. Likewise, all outputs in the layer are repeatedly become the inputs to a neuron in the subsequent layers until it reaches the output layer as many as a predefined number of hidden layers. The construction of the network is quite flexible that the number of neurons and layers can be independently varied. The use of multiple hidden layers and activation functions in their connections helps in modeling complex nonlinear relationships. The activation function can be in various forms that are differentiable such as linear, hyperbolic tangent, and sigmoid. The overall equation of one output neuron from all neurons at a current layer and an illustration of Deep Neural Network (DNN) with hidden layers are shown in Equation 5 and Figure 8, respectively.

$$h = f\left(b + \sum_{i=1}^{n} w_i x_i\right) \quad \text{Equation 5}$$

In Figure 8, $n$ is the number of nodes in the current layer. For example, $n = 3$ for the input layer if the control variables are three, such as mass, stress, and buckling load. It is essential to perform a training process before deploying the DNN for prediction. During the training, the weights and biases for all neurons are iteratively improved through an optimization process where the objective is to minimize the error in predicting the desired output for the input vector of a training dataset. Gradient descent is the most common optimization algorithm [16]. Often, the trained network does not properly respond to novel, unobserved inputs outside of the training dataset, which are called "underfitting" and "overfitting" issues. These issues can be addressed by constructing a proper size of the network and hyperparameter tuning. Furthermore, regularization techniques can be applied to the training not only to minimize the training error but also to minimize the weights themselves in the assumption that the true underlying function has a degree of smoothness. In this work, we employed the Bayesian regularization technique that seeks a balance between two objectives that minimize the training error while keeping the weights small by estimating the objective function parameters based on the Bayes rule, which leads to a better prediction than non-regularized optimizations [17]. The details on developing deep neural network architecture and training them are available in [18-21].

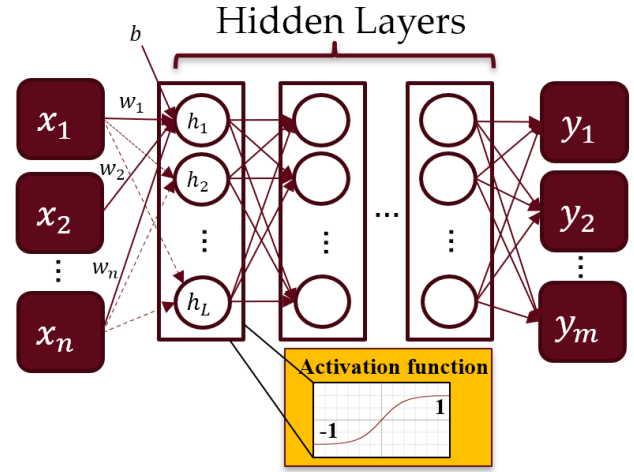

Figure 8: Deep Neural Network with Multiple Hidden Layers, n Input Units and m Output Units with L Neurons of the First Hidden Layer with Weights w and Bias b. The Activation Shown Here is Hyperbolic Tangent Function. Some Connections are Omitted for Better Visualization.

In the following section, we present the design problem. The capability of this method is demonstrated through a problem of the design of flexible disc elements in disc couplings. Deep learning is applied to the data generated through FEA simulations. Optimization is then carried out on the deep learning models for exploring optimal design solutions.

## 3. MATERIALS AND METHODS.

### 3.1 Simulation Setup

On both the designs (Design A and Design B), FEA simulations are carried out with a tangential load (accounts for torque load) and axial misalignment capability of 0.25 mm. The design variables are length (l), width (b) and thickness (t). Design A has the same width across length while Design B has a variable width as a result of circle arc. Hence, for Design B width (b) represents the width across the center of the link as shown in Figure 9.

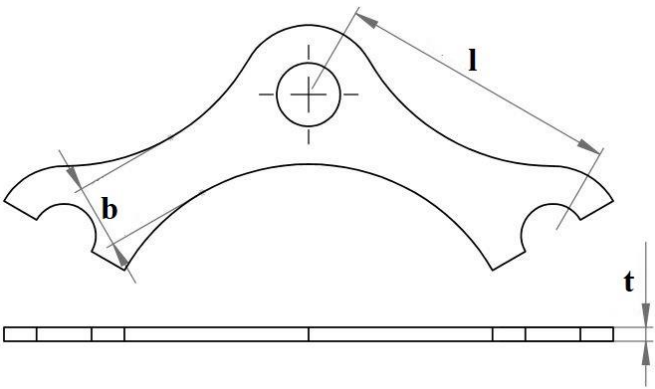

Figure 9: Design Variables in Discs Design; Length as l, Width as b, and Thickness as t

With alloy steel as a material and with previously defined loading conditions (torque and misalignment), simulations were carried out to capture the performance change as a result of change in design variables. The design variables, their bounds, and the response of interest for both the designs are tabulated in Table 2.

Table 2: Design Variables, Bounds, and Response of Interest

| Bounds | Disc Design Parameters | | | Response of interest |
|---|---|---|---|---|
| | Length mm | Width mm | Thickness mm | |
| **Low** | 24 | 3 | 0.3 | Mass |
| **Average** | 32 | 6 | 0.6 | Stress |
| **High** | 40 | 9 | 0.9 | Buckling Load |

SolidWorks is utilized in carrying out FEA simulations. From the simulation, for a given set of input variables (length, width, and thickness), the output responses of interest (mass, stress and buckling load) were recorded. All training and predictions were carried out using Matlab software. All output responses of the dataset were normalized to have zero mean and a unity standard deviation in order to make the responses are consistent in terms of magnitude so as to facilitate the training process. Also, the prediction results were scaled back to the original data range by its inverse operation. As to the neural network, the hyperbolic tangent is used as an activation function for all hidden layers and linear for the output layer because it demonstrated the best performance in terms of prediction.

### 3.2 Predicting and Optimizing Designs

The data generated are used in developing models that are to be applied in predicting and optimizing design performances. For the purpose of demonstration mass and stress are selected as performance parameters while length (l), width (b) and thickness (t) are performance predictors. The goal is to design discs with minimal stress and mass that satisfies torque and misalignment requirement while avoiding failure due to buckling. Two methods are simultaneously developed and implemented from the same data for the purpose of comparison as shown in Figure 10.

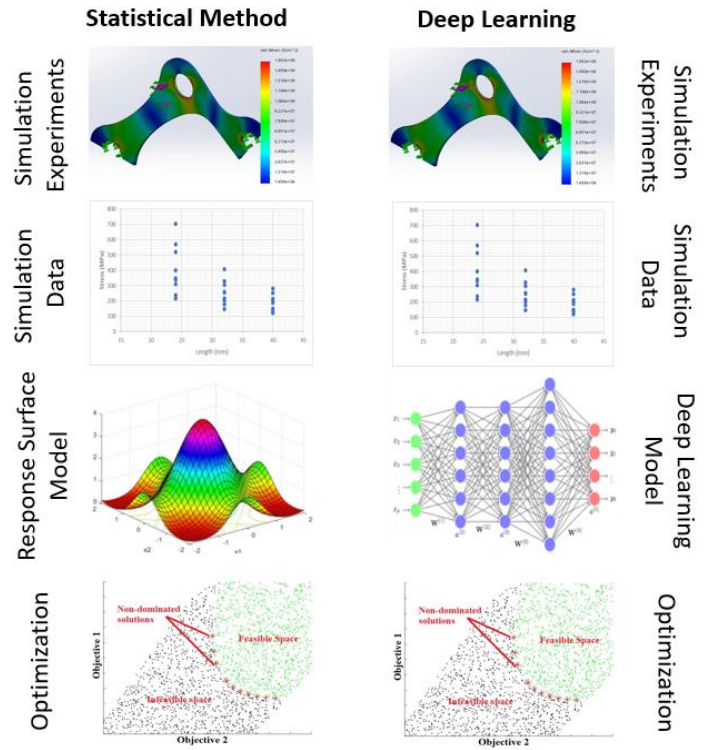

Figure 10: Two methods Illustrating Design Prediction and Optimization Strategies

Table 3: Optimization Formulation for the Two Methods

| Optimization | Statistical Method | Minimize: $f_1(x)$ and $f_2(x)$ w.r.t. $f_3(x) \geq 150$ N where, $f_1(x)$, $f_2(x)$, and $f_3(x)$ are the response surface models for mass, stress and buckling load. |
|---|---|---|

| | |
|---|---|
| Deep Learning Method | Minimize: $DNN_1(x)$ and $DNN_2(x)$<br>w.r.t. $DNN_3(x) \geq 150$ N<br>where, $DNN_1(x)$, $DNN_2(x)$, and $DNN_3(x)$ are the deep learning models for mass, stress and buckling load. |

Statistical method is applied in developing response surface models from the generated data. Equation 6, Equation 7 and, Equation 8 are the response surface models for DesignA. Equation 9, Equation 10 and, Equation 11 are the response surface models for DesignB. The response surface models are used as models for optimization. All these models have $R^2 \geq 0.93$. Similarly, deep learning method is applied on the same data for developing models with deep neural network configurations. Statistical and deep learning models are developed to predict mass, stress and buckling load as a function of length (l), width (b) and thickness (t). Using these predictive models, a method is developed to optimize design performances (mass and stress in this case) against design requirement and failure criteria.

**DESIGNA:**

$$\text{Mass} = 0.00199\, lbt - 0.00371\, bt + 0.00369 \quad \text{Equation 6}$$

$$\text{Stress} = 263.3 + 1065.3\, t^2 - 0.47\, lb - 25.1\, lt^2 \quad \text{Equation 7}$$

$$\text{Buckling load} = -0.995\, l^2 bt^3 + 2075.19\, bt^3 \quad \text{Equation 8}$$

**DESIGN B:**

$$\text{Mass} = 0.00153\, lbt + 0.01613\, lt - 0.262\, t + 0.00044 \quad \text{Equation 9}$$

$$\text{Stress} = 292.9 + 769.3\, t^2 - 5.17\, l - 17.52\, lt^2 \quad \text{Equation 10}$$

$$\text{Buckling load} = -1.47792\, l2bt^3 + 3078.22\, bt^3 \quad \text{Equation 11}$$

For design optimization in both the methods (see optimization formulation in Table 3), non-dominated sorting genetic algorithm II (NSGA-II) [22] is considered in this work. NSGA-II is a popular multi-objective optimization approach with its fast non-dominated sorting approach, a fast crowded distance estimation procedure, and simple crowded comparison operator. NSGA-II follows the general outline of a genetic algorithm with a modified mating and survival selection. The initial populations (a set of points) were selected, evaluated, and sorted based on non-domination into multiple fronts. Each individual in each front is assigned rank (fitness), values and a parameter called crowding distance that quantifies proximity of an individual to its neighbors. Low rank and larger crowding distance will give better fitness and more diversity in the population, respectively. Based on the rank and crowding distance, individuals are selected to generate offspring (the next set of points for further evaluation) through a process called a binary tournament mating selection. The evaluations results from the offspring are sorted again based on non-domination for the next selection process until it reaches the maximum number of generations (iterations). The total population is maintained with the best results throughout the procedure. A typical example of the algorithm is shown in Figure 11 for the first few generations and the last with Pareto front (also called Pareto frontier), which a set of optimal solutions usually forms a line for a problem with two objectives and a surface for three or more objectives.

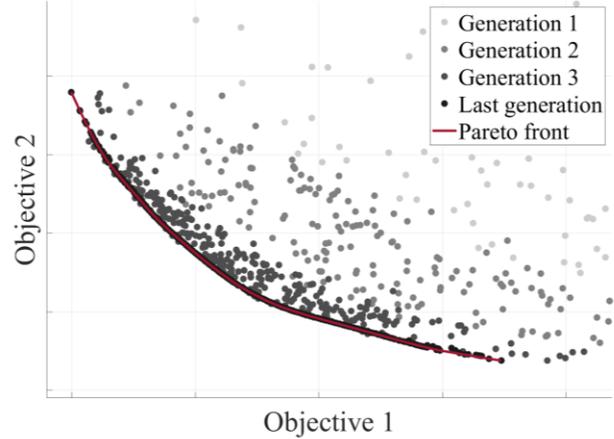

Figure 11. Example of two objectives optimization using NSGA-II for multiple generations

## 4. RESULTS AND DISCUSSION
### 4.1 Parametric Study on Network Size
The total numbers of data samples prepared for this work were 127 for Design A and 128 for Design B. In order to measure the prediction capability of the neural network, different sizes of the networks were trained with 100 training samples for both designs. On the other hand, the remaining 27 samples for Design A and 28 samples for Design B were reserved for testing so as training to test ratio to be 8 to 2. For each network, the performance was measured ten times independently with randomly drawn samples for train and test data sets with random initial values for weights and biases to account for stochastic nature wherein. As an example, the following figure (Figure 12) shows an instance of random sampling when 100 and 27 samples are drawn for training and test, respectively.

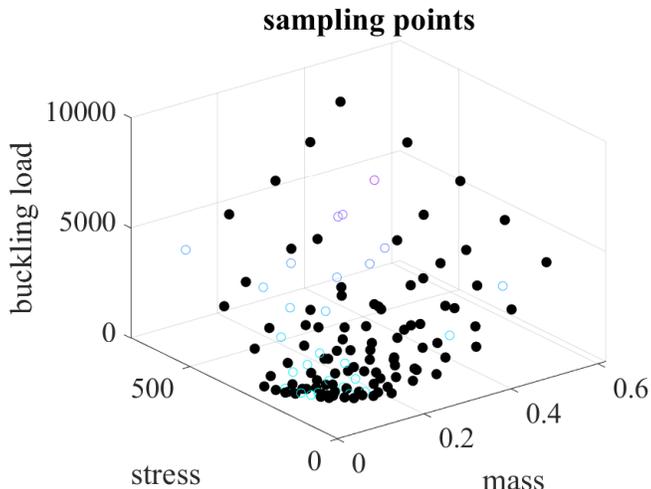

Figure 12. Example of random sampling of training and test data set for Design A. Black dots (100 points) represent training data, and open color circles (27 points) represent test data set. Every trial will be different set of training and

The error metric used here is mean absolute percent error.

$$e = \left[\frac{1}{N}\sum_j \left|\frac{\hat{y}_j - y_j}{y_j}\right|\right] \cdot 100 \qquad \text{Equation 12}$$

where $y_j$ is the ground-true response scalar value of the data sample j, $\hat{y}_j$ is the prediction value by the trained network, and $N$ is the total number of data sample. Note that this equation shows for one response, e.g., mass or stress. All of the percent errors for each response of each network were then averaged. Table 4 and Table 5 show the performance of different sizes of the network, i.e., different number of layers and neurons, in terms of prediction error for two cases when only the test data set is used and both test and training data set are used altogether.

Several observations can be made. First, increasing the number of neurons generally improves the accuracy but the differences are marginal if 20 or more neurons are used. Second, adding more layers does not guarantee better accuracy as seen in the inconsistent results of the three layers. Third, overall low standard deviations (less than one percent) indicate that trained networks are stable and robust to the random initialization and random selections of samples for the training dataset. Finally, the overall error when using all data is not much different when using only test data even with the increased number of parameters, indicating the models are well-generalized with the help of the Bayesian regularization. In other words, it is expected to correctly respond to novel unseen data. Therefore, we can safely choose any networks between 1 to 2 layers with 20-40 neurons within. In this work, we chose 2 hidden layers that have 20 neurons (2×20) because it is the simplest model that gives better or comparable predictions.

Table 4: Parametric Study On Network Size For Design A (Percent Error With Std. Deviation). "All" Data Include Training and Test Data.

| Design A | | Number of Neurons | | | |
|---|---|---|---|---|---|
| | | 10 | 20 | 30 | 40 |
| Number of Layers | 1 Test | 3.7±1.8 | 2.2±0.4 | 1.9±0.7 | 2.1±0.6 |
| | 1 All | 3.1±0.8 | 1.2±0.1 | 0.9±0.3 | 0.9±0.3 |
| | 2 Test | 1.9±0.5 | **1.7±0.5** | 1.6±0.3 | 1.4±0.3 |
| | 2 All | 0.9±0.2 | **0.6±0.1** | 0.5±0.1 | 0.5±0.1 |
| | 3 Test | 32.3±67.9 | 1.6±0.4 | 11.1±30.5 | 1.5±0.3 |
| | 3 All | 33.5±69.2 | 0.6±0.1 | 15.7±47.9 | 0.5±0.1 |

Table 5: Parametric Study On Network Size For Design B (Percent Error With Std. Deviation). "All" Data Include Training and Test Data.

| Design B | | Number of Neurons | | | |
|---|---|---|---|---|---|
| | | 10 | 20 | 30 | 40 |
| Number of Layers | 1 Test | 3.4±1.6 | 2.5±0.9 | 2.0±0.7 | 2.0±0.8 |
| | 1 All | 2.8±0.6 | 1.3±0.3 | 0.8±0.2 | 0.8±0.2 |
| | 2 Test | 2.1±0.7 | **1.6±0.5** | 1.6±0.6 | 1.5±0.3 |
| | 2 All | 0.9±0.3 | **0.6±0.1** | 0.5±0.2 | 0.5±0.1 |
| | 3 Teti | 32.1±68.1 | 1.4±0.2 | 1.5±0.5 | 1.4±0.3 |
| | 3 All | 33.5±69.4 | 0.5±0.1 | 0.5±0.2 | 0.5±0.1 |

**4.2 Parametric Study on Different Size of Training Dataset**

In this section, we studied the size effect of the training dataset. Different numbers of training set are used for training while the network is fixed as 2×20 as chosen in the previous section. The change in the size of the training data resulted in the different sizes of test data. The same as the study in the previous section, 10 random trials were averaged for statistical analysis. Also, predictions were performed for test-only data and test-and-training data to show how the models are well-generalized. As seen in the tables (Table 6 and Table 7), overall errors go down as more training data are added, but the differences converge. One thing to note is that the standard deviation of the Design B test prediction result with 120 training data is higher than that of smaller number of training data, indicating that training data were overfitted.

Table 6: Parametric Study on Training Data Size for Design A (Percent Error with Std. Deviation). "All" Data Include Training and Test Data. Note: The Number of Test data = 127 Total Data – The Number of Training Data.

| DesignA | Number of Training Data | | | | |
|---|---|---|---|---|---|
| | 40 | 60 | 80 | 100 | 120 |
| Test | 5.6±2.8 | 3.5±1.2 | 2.2±0.4 | **1.7±0.5** | 1.4±0.2 |
| All | 4.1±1.9 | 2.1±0.7 | 1.0±0.2 | **0.6±0.1** | 0.3±0.1 |

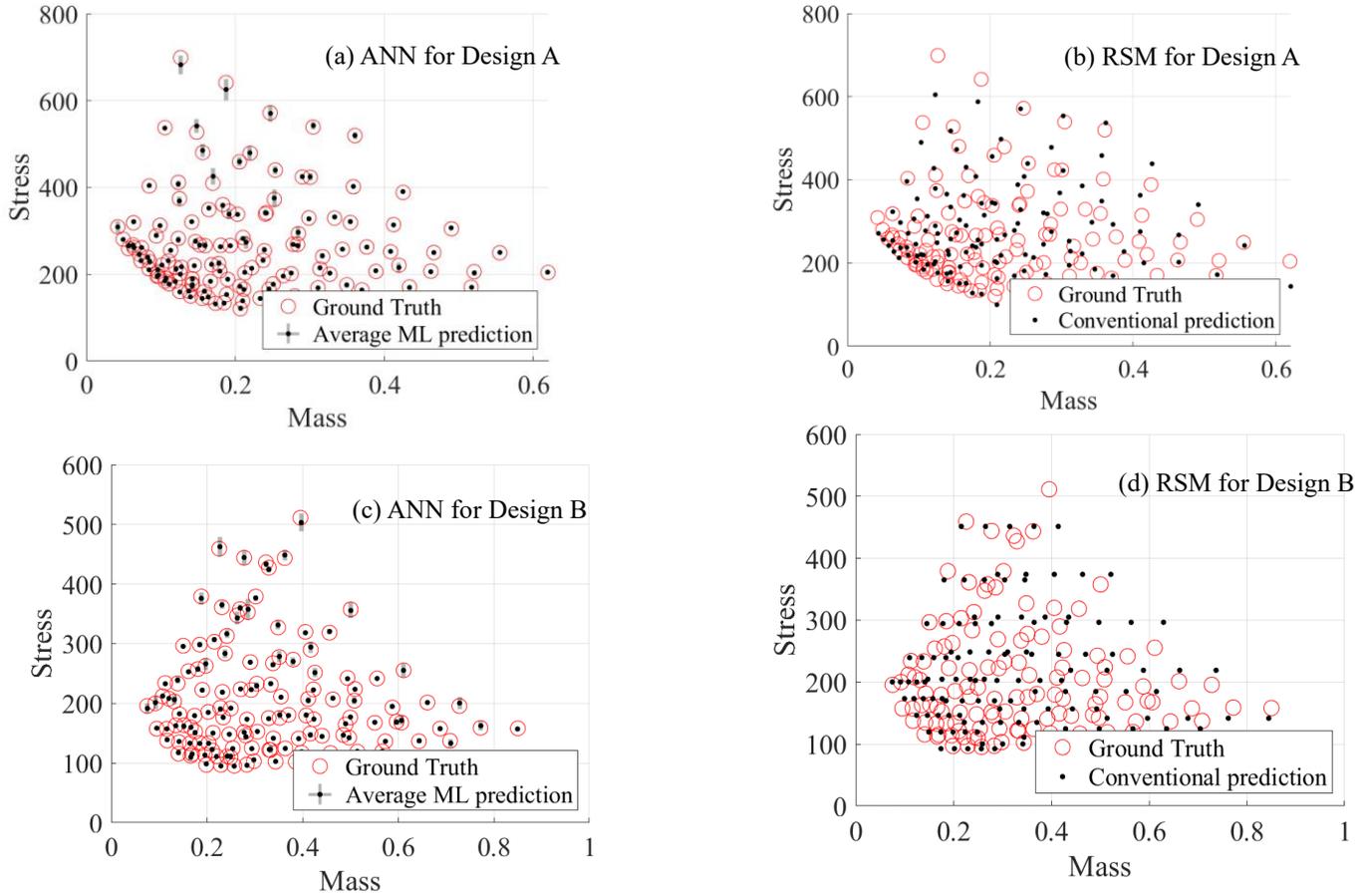

Figure 13: Prediction comparison with Response Surface Model (RSM). (a) 2×20 ANN and (b) RSM for Design A, (c) 2×20 ANN and (d) for Design B. Circles represent ground truth data and dots predictions. For ANN, the crossbar represents the standard deviation of 10 different networks and data compositions.

Table 7: Parametric Study On Training Data Size For Design B (Percent Error With Std. Deviation). "All" Data Include Training and Test Data. Note: The Number of Test Data = 128 Total Data – The Number of Training Data.

| DesignB | Number of Training Data | | | | |
|---|---|---|---|---|---|
|  | 40 | 60 | 80 | 100 | 120 |
| Test | 4.6±1.4 | 3.4±1.0 | 2.1±0.5 | **1.6±0.5** | 1.4±0.7 |
| All | 3.3±0.9 | 2.0±0.6 | 1.0±0.2 | **0.6±0.1** | 0.3±0.1 |

**Comparison with conventional surrogate modeling:**
In Figure 13, we show the visual comparison of the prediction results using ANN, specifically 2×20 architecture, and the response surface model (RSM) for Design A and B when all training and test data set are utilized. In the figure, stress as a function of mass is plotted. Unlike RSM, the ANN predictions are the averaged results, so they are plotted with crossbars for standard deviation. Both of the approaches generally agree with the ground truth data, but the ANN predicts better for the extreme cases where data are fewer. Also, overall narrow crossbars indicate that with any random initialization of parameters and randomly selected training data the developed networks reliably predict the results. Also, it is shown that the same 2×20 architecture works for both Design A and B equally well. The overall results, along with the results of the parametric study, give designers confidence that they can choose a reasonably well-trained model without having to tune the model too much for a particular problem.

Finally, we performed design optimization through the genetic algorithm, NSGA-II. A Matlab implementation of NSGA-II was utilized for the simulations [23]. The original codes were modified to seamlessly work with our DNN models. The codes were changed to have more vectorization, which improved the computation performance by factor of two. Stress and mass were simultaneously optimized for designs that have an ability to withstand at least buckling load equal to 150 N. 150 N is selected to ensure that each feasible disc design has the capability to transmit at least 30 Nm of torque (derived using Equation 4 and variable bounds shown in Table 2). The simulations were performed with 500 populations for 300 generations (iterations).The optimization results from genetic algorithms are compared in Figure 14, Table 8, and Table 9. The solutions of the final generations are plotted with two extreme solutions and one optimum solution that is the point in case no importance for each objective is considered. Specifically, the "optimum" solution is one that has the minimum distance between the zero-reference point and all possible optimal solutions for mass and stress in the Pareto front. For the distance measure, all solutions were normalized with zero mean and unity standard deviation.

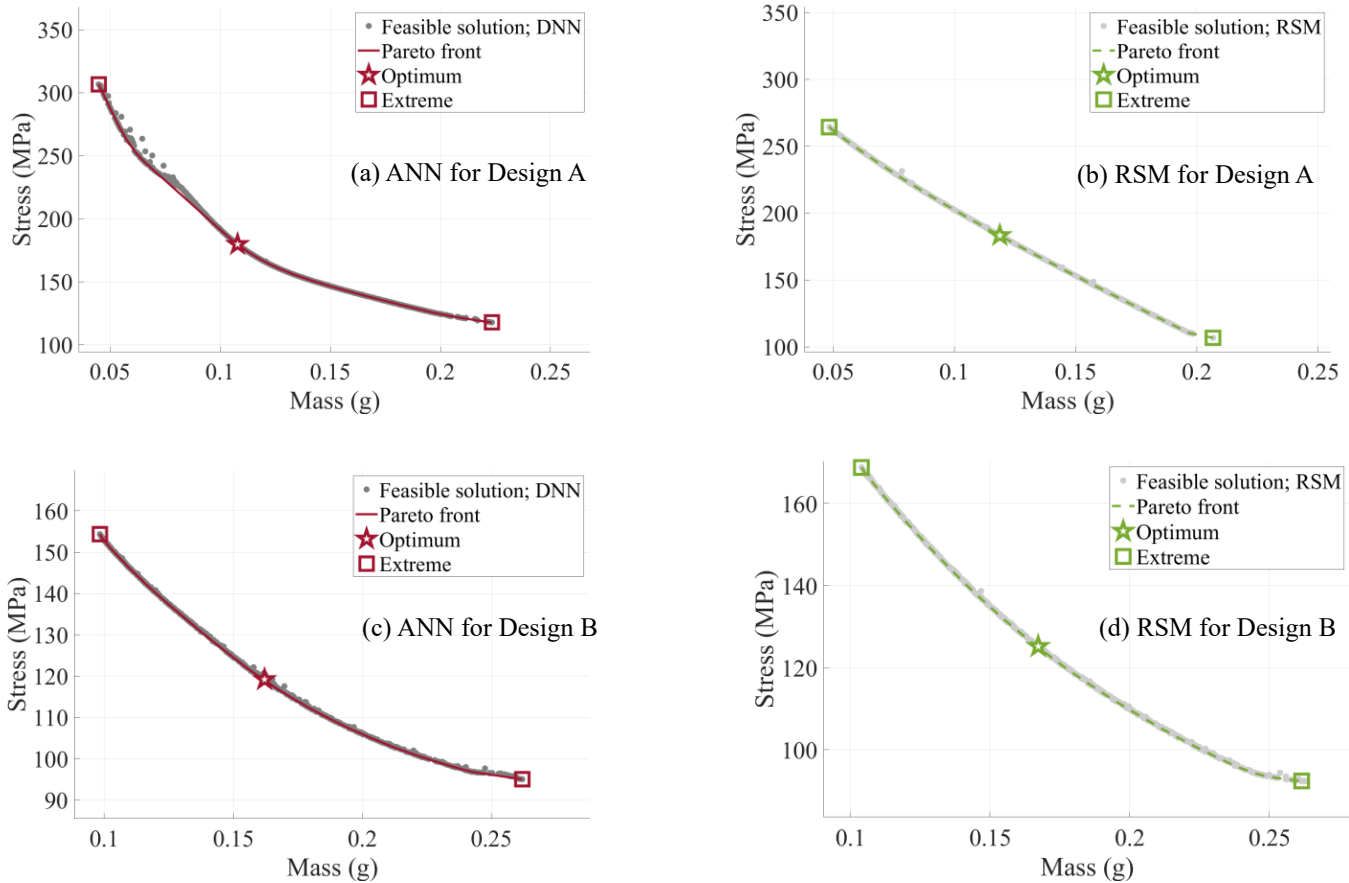

Figure 14. Comparison of Optimization Results. Optimization Objectives are Stress and Mass While Buckling Load is a Constraint. Star Markers Represent Optimum Points if Stress and Mass Objectives are Considered to Have Equal Importance.

In Table 8, it can be seen that statistical method optimizes mass for Design A to 0.05 g while deep learning method optimizes mass for Design A to 0.04 g. Similarly, statistical method optimizes stress for Design A to 106.72 MPa while deep learning method optimizes stress for Design A to 118.68 MPa. In Table 9, it can be seen that statistical method optimizes mass for Design B to 0.10 g and deep learning method also optimizes mass for Design B to 0.10 g. Similarly, statistical method optimizes stress for Design A to 92.39 MPa while deep learning method optimizes stress for Design A to 95.05 MPa.

Table 8: Optimization results for Design A. Length, Width, and Thickness are in Millimeters, Mass in g, and Stress in MPa. "Optimum" Point if Stress and Mass Objectives are Considered to Have Equal Importance.

| Design A | | Design Parameter | | | Objective | |
|---|---|---|---|---|---|---|
| | | Length | Width | Thickness | Mass | Stress |
| DNN | Minimal Mass | 24.00 | 3.00 | 0.32 | 0.04 | 306.60 |
| | Minimal Stress | 39.42 | 9.00 | 0.31 | 0.22 | 118.68 |
| | "Optimum" | 33.06 | 5.65 | 0.30 | 0.11 | 179.64 |
| RSM | Minimal Mass | 24.00 | 3.00 | 0.30 | 0.05 | 264.30 |
| | Minimal Stress | 38.99 | 9.00 | 0.31 | 0.21 | 106.72 |
| | "Optimum" | 34.25 | 6.12 | 0.30 | 0.12 | 183.32 |

Table 9: Optimization Results for Design B. Length, Width, and Thickness are in Millimeters, Mass in g, and Stress in MPa. "Optimum" Point if Stress and Mass Objectives are Considered to Have Equal Importance.

| Design B | | Design Parameter | | | Objective | |
|---|---|---|---|---|---|---|
| | | Length | Width | Thickness | Mass | Stress |
| DNN | Minimal Mass | 28.57 | 3.00 | 0.30 | 0.10 | 154.34 |
| | Minimal Stress | 40.00 | 7.76 | 0.30 | 0.26 | 95.05 |
| | "Optimum" | 35.59 | 3.88 | 0.31 | 0.16 | 119.19 |
| RSM | Minimal Mass | 28.66 | 3.00 | 0.30 | 0.10 | 168.83 |
| | Minimal Stress | 40.00 | 7.75 | 0.30 | 0.26 | 92.39 |
| | "Optimum" | 35.34 | 4.51 | 0.30 | 0.17 | 125.21 |

In Table 8, it can be seen that statistical method optimizes mass for Design A to 0.05 g while deep learning method optimizes mass for Design A to 0.04 g. Similarly, statistical method optimizes stress for Design A to 106.72 MPa while deep learning method optimizes stress for Design A to 118.68 MPa. In Table 9, it can be seen that statistical method optimizes mass for Design B to 0.10 g and deep learning method also optimizes mass for Design B to 0.10 g. Similarly, statistical method optimizes stress for Design A to 92.39 MPa while deep learning method optimizes stress for Design A to 95.05 MPa.

## 5. CONCLUSION

In recent years, machine learning-based predictive models have been used to generate solutions to complex engineering problems. In this paper, we couple the machine learning developed predictive models with a genetic algorithm to identify optimal design solutions of flexible disc elements. The data generated from the simulation are used to train the machine learning models. A genetic algorithm utilized these trained models to navigate the design space and identify optimal design solutions. To demonstrate the efficacy of the method, we compare it with results obtained from statistical method.

Another important aspect of this paper is to enable disc coupling designers to leverage machine learning methods in identifying optimal design solutions. First, we establish the necessary foundation required for designing flexible discs for disc couplings. Starting with two different discs design, we showcase how we navigate through the design space and identify solutions that are optimal. Our intention is to establish value in the proposed method as a generic method that can be applied in a variety of engineering design problems.

## ACKNOWLEDGEMENTS

We thank the Center for Advanced Vehicular Systems (CAVS) at Mississippi State University for supporting this effort.